\pdfoutput=1

\documentclass[11pt]{article}

\usepackage{ACL2023}

\usepackage{times}
\usepackage{latexsym}

\usepackage[T1]{fontenc}

\usepackage[utf8]{inputenc}

\usepackage{microtype}

\usepackage{inconsolata}

\usepackage{graphicx}
\usepackage{amsmath}
\usepackage{multirow}
\usepackage{CJKutf8}
\usepackage{color}
\usepackage{diagbox}
\usepackage{enumitem}
\setlist[itemize]{leftmargin=*}

\title{I run as fast as a rabbit, can you? A Multilingual Simile Dialogue Dataset}

\author{Longxuan Ma$^1$ \and Weinan Zhang$^1$\thanks{*Corresponding author} \and Shuhan Zhou$^{1,2}$ \\
 \and {\bf Churui Sun}$^3$ \and {\bf Changxin Ke}$^3$ \and {\bf Ting Liu}$^1$\\
$^1$ Research Center for Social Computing and Information Retrieval, \\ Harbin Institute of Technology\\ 
$^1$ \texttt{lxma,wnzhang,shzhou,tliu@ir.hit.edu.cn} \\
$^2$   School of Information Science, Beijing Language and Culture University \\
$^3$  School of Computer Science, Harbin Institute of Technology\\
$^3$  \texttt{sunchurui@hit.edu.cn, cxke@stu.hit.edu.cn} \\
}

\begin{document}
\maketitle
\begin{abstract}
A simile is a figure of speech that compares two different things (called the tenor and the vehicle) via shared properties. The tenor and the vehicle are usually connected with comparator words such as "like" or "as". The simile phenomena are unique and complex in a real-life dialogue scene where the tenor and the vehicle can be verbal phrases or sentences, mentioned by different speakers, exist in different sentences, or occur in reversed order. However, the current simile research usually focuses on similes in a triplet tuple (tenor, property, vehicle) or a single sentence where the tenor and vehicle are usually entities or noun phrases, which could not reflect complex simile phenomena in real scenarios. In this paper, we propose a novel and high-quality multilingual simile dialogue (MSD) dataset to facilitate the study of complex simile phenomena. The MSD is the largest manually annotated simile data ($\sim$20K) and it contains both English and Chinese data. Meanwhile, the MSD data can also be used on dialogue tasks to test the ability of dialogue systems when using similes. We design 3 simile tasks (recognition, interpretation, and generation) and 2 dialogue tasks (retrieval and generation) with MSD. For each task, we provide experimental results from strong pre-trained or state-of-the-art models. The experiments demonstrate the challenge of MSD and we will release the data/code on GitHub.
\end{abstract}

\section{Introduction}
Simile plays an important role in human language to make utterances more vivid, interesting, and graspable \cite{DBLP:conf/aaai/ZhangCXGLWC21,DBLP:conf/acl/HeCLXX22} and is an increasingly studied phenomenon in computational linguistics \cite{DBLP:journals/taslp/SongGFLL21,DBLP:conf/acl/HeCLXX22}. A simile is a figure of speech that compares two things from different categories (called the tenor and the vehicle) via shared properties \cite{AnthonyPaul1970}. A tenor and a vehicle are usually connected with comparator words such as "like" or "as". For example, in the first example of Table \ref{examples}, the tenor is "The boy", the vehicle is "a rabbit", the event is "run", the comparator is "as ... as" and the shared property is "fast".

\begin{table}
\footnotesize
\centering
\begin{tabular}{l|l|c}
\hline
 & \textbf{Examples} & \textbf{Simile}\\
\hline
1&\underline{The boy} runs as fast as \textit{a rabbit}. & Yes\\
\hline
2&The girl looks like her mother. & No \\
\hline
&A: Look \underline{that fireman} over the street. & \\
3&B: Wow, he is so strong. & Yes\\
&A: I agree, strong as \textit{a bull}. &\\
\hline
4&A: Like \textit{a monster}, right? &Yes\\
&B: Yes, \underline{that man} is really rude.&\\
\hline
5&A: \underline{Arguing with parents} is not wise. &Yes\\
&B: It is like \textit{throwing an egg at a rock}.&\\
\hline
6&A: \underline{He walks into the crowd} and disappears. &Yes\\
&B: It is like \textit{a fish swims into the ocean}.&\\
\hline
\end{tabular}
\caption{Examples to illustrate simile. The underline font represents \textbf{tenors}. The italic font means \textit{vehicles}. A and B are different Speakers.}
\vspace{-0.5cm}
\label{examples}
\end{table}

The current simile research usually focuses on the simile in a triplet (tenor, shared property, vehicle) \cite{DBLP:journals/taslp/SongGFLL21} or a single sentence \cite{DBLP:conf/acl-figlang/BizzoniL18,DBLP:conf/emnlp/LiuHSFLH18,DBLP:conf/coling/LiLG22a}. For example, the simile recognition \cite{DBLP:conf/eacl/BirkeS06,DBLP:conf/emnlp/LiuHSFLH18} task is judging whether a sentence contains a simile, such as distinguishing which of the first and second examples in Table \ref{examples} contains a simile. However, a simile in a triplet or a single sentence is not enough to reflect the complex simile phenomena in the real scenario. In this paper, we study similes in real-life dialogue where a tenor and a vehicle can be mentioned by different speakers, exist in different sentences, or occur in reversed order. The third example in Table \ref{examples} shows a simile dialogue where the tenor "That fireman" and the vehicle "a bull" are in different utterances. The fourth example in Table \ref{examples} shows a simile where the tenor and the vehicle are mentioned by different speakers and the vehicle occurs before the tenor. What is more, different from previous research where the tenor and vehicle are usually single entities \cite{DBLP:journals/taslp/SongGFLL21} or simple nominal phrases \cite{DBLP:conf/acl-figlang/BizzoniL18}, \textit{the tenor and vehicle in a real-life dialogue may be a verbal phrase or a long sentence}. A verbal phrase can function as the subject or object of a verb, such as the fifth example in Table \ref{examples}. A sentence is a set of words expressing a statement, a question, or an order, usually containing a subject and a verb. The sixth example in Table \ref{examples} shows sentences as the tenor and vehicle. The verbal phrase and sentences can convey richer content and emotions, making the real-life dialogue more vivid and interesting. Studying these complex simile phenomena in a dialogue scenario needs to consider both the dialogue context and the various forms of the tenor and vehicle, and will lead the simile research to a brand new level. However, similes in real-life dialogue scenarios have not been studied by previous research so there are no public benchmarks available nowadays.

To facilitate the simile study, we release a human-annotated, high-quality simile dialogue dataset, which contains both English and Chinese data. The complex simile phenomena in real-life dialogue scenarios not only bring more difficulties to traditional simile tasks such as recognition, interpretation \cite{DBLP:journals/eaai/SuTC16}, and generation \cite{DBLP:conf/coling/LiLG22a} but also raise challenges for dialogue research, e.g. generation and retrieval tasks. To address the simile dialogue tasks, dialogue models need to understand the simile relations between entities/phrases/sentences. Our contributions are:
\begin{itemize}
\item To the best of our knowledge, we are the first to study the simile phenomenon in dialogue and propose a high-quality multi-lingual simile dialogue (MSD) dataset to assist both the simile and dialogue research.
\item There are 5 tasks with the proposed MSD dataset. For simile research, we design the dialogue simile recognition/interpretation/generation tasks. For dialogue research, we design the response retrieval and generation tasks.
\item We verify how strong pre-trained models and the state-of-the-art simile models perform on the 5 tasks we designed. Experimental results reveal that simile in dialogue is a difficult task and requires further study. Our code and data will be released on GitHub\footnote{https://github.com/malongxuan/MSD}.
\end{itemize}

\begin{table}
\footnotesize
\centering
\begin{tabular}{l|l}
\hline
\textbf{Metaphor Category} & \textbf{Example}\\
\hline
Noun phrase & \underline{The nurse} is \textit{an angel}. \\
\hline
Adjective   & These words are cold. \\
                 & The soldier had a warm heart.\\
\hline
Verbal   & The process was killed. \\
              & They plant the seeds of change. \\
\hline
Adverb-Verb  & He speak fluidly. \\
\hline
Verbal phrase & \underline{Taking care of pets} is like\\
& \textit{raising children}.\\
\hline
Sentence  & \underline{I rushed to the terminal} like\\
& \textit{a cheetah chasing its prey}.\\
\hline
\end{tabular}
\caption{Different metaphor categories. The underline font represents \textbf{tenors}. The italic font means \textit{vehicles}. The similes in our MSD data cover Noun phrases, Verbal phrases, and Sentence categories. The two examples in Adjective show two different Adjective-Noun modes. The two examples in Verbal are Subject-Verb and Subject-Verb-Object modes. }
\vspace{-0.5cm}
\label{metaphor}
\end{table}

\section{Related Work}

\subsection{Simile and Metaphor}
The simile is a kind of metaphor that is frequently used in human languages to make utterances more vivid and graspable \cite{DBLP:conf/emnlp/NiculaeD14} and expresses human sentiments \cite{DBLP:conf/wism/LiKZCT12,DBLP:conf/starsem/MohammadST16}. Previous researchers defined different metaphor categories. We present examples for these categories in the first four lines of Table \ref{metaphor}. For example, \citet{DBLP:conf/acl-figlang/BizzoniL18} categorized metaphor into Noun phrases, Adjectives, Verbs, and Multi-word; \citet{DBLP:conf/coling/LiLG22a} categorized metaphor into Nominal, Verbal (Subject-Verb-Object), Adjective-Noun, and Adverb-Verb. Previous work usually denoted the Noun phrase metaphor as a simile \cite{DBLP:conf/coling/LiLG22a,DBLP:conf/acl/HeCLXX22,DBLP:conf/acl/ChenCZPCZXCS22}. \textit{Following previous work, we also categorize Noun phrase metaphor as a simile. Meanwhile, we extend the tenor and vehicle to verbal phrases and sentences according to the simile phenomena in dialogue.} The examples of verbal phrases and sentences in simile are shown in the last two lines of Table \ref{metaphor}.

\subsection{Tasks in Metaphor/Simile}

The tasks in metaphor are also suitable for simile, such as recognition \cite{DBLP:conf/eacl/BirkeS06,DBLP:conf/emnlp/LiuHSFLH18}, interpretation \cite{DBLP:journals/eaai/SuTC16}, and generation \cite{DBLP:conf/coling/LiLG22a}. The recognition task is also called identification \cite{GerardSteen2010,DBLP:conf/coling/LiLG22a} or detection \cite{DBLP:conf/acl/TsvetkovBGND14,DBLP:conf/lrec/MohlerBRT16}, which aims to identify whether a given phrase or sentence contains a metaphor/simile. The interpretation is also called explanation \cite{DBLP:conf/emnlp/LiuHSFLH18} which usually assigns an appropriate interpretation to a metaphorical expression \cite{DBLP:conf/acl-figlang/BizzoniL18} or infers the shared properties of the tenor and the vehicle \cite{DBLP:journals/taslp/SongGFLL21,DBLP:conf/acl/HeCLXX22,DBLP:conf/acl/ChenCZPCZXCS22}. The generation task also has different forms. For example, when giving an input tenor, it can generate a simile sentence conditioned on the input tenor \cite{DBLP:conf/coling/LiLG22a}; when giving both the tenor and the shared property in simile, it can generate the vehicle \cite{DBLP:journals/taslp/SongGFLL21,DBLP:conf/acl/ChenCZPCZXCS22}; when providing a literal sentence, it can generate a metaphoric sentence which paraphrases that input \cite{DBLP:conf/emnlp/ChakrabartyMP20,DBLP:conf/conll/StoweBG21}, or generating a specific simile according to the location where the simile interpolation should happen \cite{DBLP:conf/aaai/ZhangCXGLWC21}. In this paper, we also define recognition, interpretation, and generation tasks. \textit{However, different from previous work that only focused on similes in a triplet tuple or a sentence, we investigate a more challenging scenario where the simile happens in a multi-turn dialogue.}

\begin{table}
\footnotesize
\centering
\begin{tabular}{c|c|c|c|c|c}
\hline
\textbf{Dataset} & \textbf{Lan.} & \textbf{Form}  & \textbf{Task}  & \textbf{Size} & \textbf{Man.}\\
\hline
CM & Ch &sentence& I &  85 & Yes \\
SRC & Ch & sentence & R & 11,337 & Yes\\
CMC & Ch & sentence  & G  & 11,581 & Yes\\
MCP & En & sentence & I &  1,633 & Yes \\ 
\hline
SLS & En & sentence & G & 87K & No \\
WPS & Ch & sentence  & G  & 5M  & No\\
\hline
Ours & Ch/En & Dialogue & R/I/G & 19,565 & Yes\\
\hline
\end{tabular}
\caption{Survey of existing simile datasets. "Lan."/ "Ch"/ "En"/ "R"/ "I"/ "G"/ "Man." is short for "Language"/ "Chinese"/ "English"/ "recognition"/ "interpretation"/ "generation"/ "manual", respectively.}
\vspace{-0.5cm}
\label{simile-datasets}
\end{table}

\subsection{Survey of Simile Datasets}

Table \ref{simile-datasets} shows the comparison between our MSD dataset with the existing simile datasets. \citet{DBLP:journals/eaai/SuTC16} constructed a small Chinese Metaphor (CM) data with 85 nominal and 35 verbal metaphors for the interpretation task. \citet{DBLP:conf/emnlp/LiuHSFLH18} introduced Simile Recognition in Chinese (SRC) data containing sentences with a special comparator \begin{CJK}{UTF8}{gbsn}像\end{CJK} (like). The Chinese Nominal Metaphor Corpus (CMC) \cite{DBLP:conf/coling/LiLG22a} data merges other Chinese metaphor datasets \cite{DBLP:conf/emnlp/LiuHSFLH18} for simile generation. \citet{DBLP:conf/acl/HeCLXX22} proposed a simile property probing task and constructed Multi-choice Probing (MCP) datasets. \citet{DBLP:conf/emnlp/ChakrabartyMP20} collected Reddit comments containing similes and then auto-constructed a parallel simile corpus with a pre-trained model powered by commonsense knowledge \cite{DBLP:conf/acl/BosselutRSMCC19}. However, their Self-labeled Similes (SLS) dataset is limited to a “like a” pattern which appears only at the end of a sentence. \citet{DBLP:conf/aaai/ZhangCXGLWC21} introduced the Writing Polishment with Similes (WPS) dataset where models need to locate the simile position in a sentence and then generate a simile in that position. The SLS and WPS are much larger than other existing data but they are not manually annotated. \textit{Our MSD data is extracted from more than 166M dialogue data (shown in Table \ref{collected-data}). It is the first multi-lingual simile dialogue data and the largest \textbf{manually annotated simile data} so far. What's more, benefiting from the strict annotation schedule, the MSD contains necessary simile components so that it can be used for simile recognition/interpretation/generation simultaneously.}

\section{Multilingual Simile Dialogue Dataset}
In this section, we introduce the collection, annotation, and statistics of our MSD data.

\begin{figure*}[t]
\centering
\includegraphics[width=\linewidth]{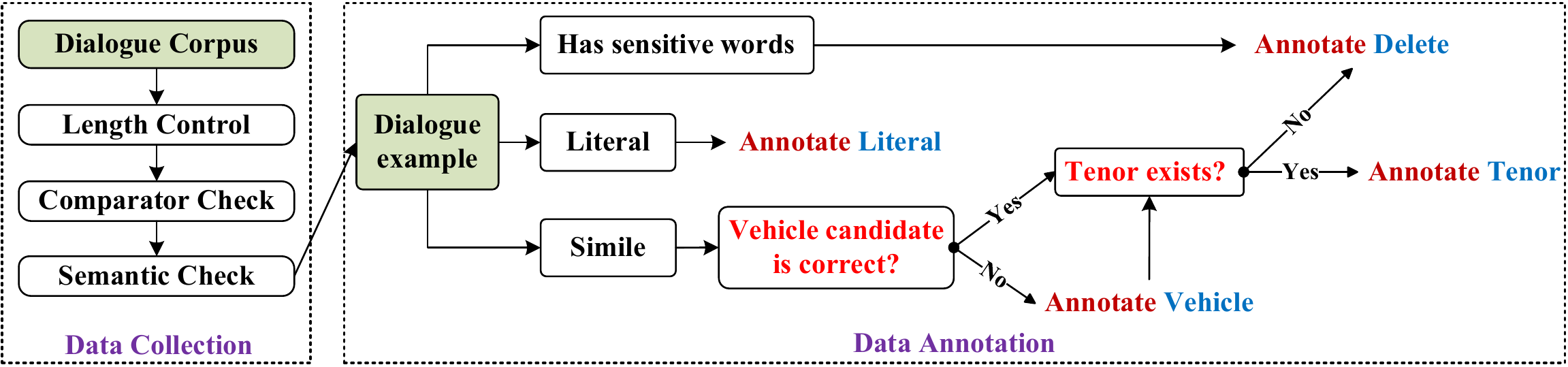}%
\caption{The data collection and annotation process.}
\vspace{-0.5cm}
\label{annotation}
\end{figure*}

\subsection{Data Collection}

Since we aim to extract the simile in a real-life dialogue, we adopt the existing open-domain dialogue corpus collected from social platforms such as Reddit.com and Weibo.com. For English similes, we use the 3 turns version Reddit Dialogue dataset \cite{DBLP:journals/corr/abs-1811-01063} which contains more than 15 million dialogues. For Chinese similes, we use two datasets: PchatbotW and LCCC. The PchatbotW \cite{DBLP:conf/sigir/QianLZGMZLDW21} is the largest dialogue data we can find and contains 139 million 2 turns dialogues from Weibo. The LCCC \cite{DBLP:conf/nlpcc/WangKZHJZH20} is also from Weibo and contains 12 million 2 or 3 turns dialogues. We treat the last utterance in a dialogue as a response and the utterances in front of the response as a dialogue context. We extract dialogues from these large-scale datasets with a rigorous data collection pipeline, which is built based on a set of rules we will introduce in this section. Notice that we do not make any changes to the original dialogue data and only extract those dialogues with comparators in the response.

\begin{table}
\footnotesize
\centering
\begin{tabular}{c|c|c|c|c}
\hline
\multirow{2}{*}{\textbf{Dataset}} & \multicolumn{4}{c}{\textbf{Dialogue examples}} \\
\cline{2-5}
& Original & Coarse & Fine & Final \\
\hline
LCCC          & 12M   & 20K & 4K & 1,214\\ 
PchatbotW & 139M & 1M & 82K &12,830\\ 
 \hline
Reddit-dialogue         & 15M   & 71K & 32K & 8,510\\
\hline
\end{tabular}
\caption{Statistics of the dialogue datasets we collected.}
\vspace{-0.5cm}
\label{collected-data}
\end{table}

In the first step, we select the dialogue examples where the responses contain comparators such as "\begin{CJK}{UTF8}{gbsn}像...一样\end{CJK}"/"like"/"as...as"\footnote{The full Chinese comparators are listed in Appendix \ref{comparators-chinese}}. We only select dialogue examples with context lengths between 15 and 30 words so that the dialogue context is both informative and not too long for the annotators to read. These examples are denoted by \textit{the coarse version} of the simile dialogue data and the statistics are shown in Table \ref{collected-data}.

In the second step, we use machine translation\footnote{We use https://ai.youdao.com/ to conduct the translation.} to ensure that a sentence contains a comparator. We only reserve the dialogue examples that still contain comparator when they are translated into another language. For example, an English simile candidate sentence "I run as fast as a rabbit" contains a comparator "as...as". When translating it into Chinese, this sentence is "\begin{CJK}{UTF8}{gbsn}我跑得像兔子一样快\end{CJK}" and still contains a comparator "\begin{CJK}{UTF8}{gbsn}像\end{CJK}"(like). After the machine translation checking, we got \textit{the fine version} of the simile dialogue candidates. The fine version needs further improvement since the candidate tenor/vehicle connected by the comparator is not always a simile. For example, the sentence "The Poodle is as tall as a Corgi" is not a simile since the sentence compares the height of two different kinds of dogs. So we conduct a third step to further remove examples that are not similes.

In the third step, we adopt a semantic dependency tool\footnote{https://stanfordnlp.github.io/CoreNLP} to locate the candidate tenor/vehicle, then we compute the similarity between them to retain the examples with low similarity so that the remaining candidate tenor/vehicle are from different categories. The similarity is computed with dense representations of the candidate tenor/vehicle from BERT \cite{DBLP:conf/naacl/DevlinCLT19}. After the above pipeline, we obtain \textit{the final version} of simile dialogue data for annotation. The statistics of the fine/final version we obtained are also shown in Table \ref{collected-data}.

\subsection{Data Annotation}
We recruited 7 students majoring in English for annotating the English data and recruit other 6 well-educated native speakers (graduate students) for annotating the Chinese data. We randomly select 100 examples in the final version, finding that the vehicle candidates we extracted have an acceptable accuracy (above 80\%). However, the accuracy of the tenor candidate is not good (below 60\%). Hence, we provided annotators with "dialogue context", "response", "comparator", and "vehicle candidate" for each dialogue. We use the annotation tool proposed by \citet{DBLP:conf/acl/YangZLL18} to simplify the operation so that the annotators can use a mouse and a few shortcuts on the keyboard to annotate. 

There are some difficulties when annotating similes in the dialogue scenario apart from the fact that the tenor may exist in different sentences or occur after the vehicle. For example, the tenor may not exist in the dialogue even if the response is a simile. We ask the annotators to delete these examples. There are other situations that a dialogue that contains commonly used phrases or slang that makes the dialogue seem like a simile but not. For instance, "make like a tree" is not a simile but slang means "leave". Besides, English words usually have different meanings. For example, according to the Oxford Dictionary, the word "body" means "the whole physical structure of a human or an animal" as well as "a group of people who work or act together, often for an official purpose". So the sentence "This association is like the body that represents its members." is not a simile. Furthermore, there are many abbreviations used on social platforms such as FTW (for the win) and OP (original poster). These difficult linguistic phenomena require the annotators to have a good understanding of the dialogue context so that they could determine whether a response contains a simile.


We conduct preliminary training for the recruited annotators so that they are aware of the professional standards. We ask the annotators to first check whether the response in this dialogue example contains a simile. The example will be annotated "Literal" if the response is not a simile. Otherwise, they should check whether the vehicle candidate in the response is correct. They need to annotate the correct vehicle (can be word/phrase/sentence) if the candidate is not accurate. If the candidate vehicle is correct, they can annotate the tenor (can be word/phrase/sentence) if it exists. We present the annotation schedule in Figure \ref{annotation}. Our annotation schedule ensures that the tenor and vehicle are in the data.

\textbf{Quality Evaluation.} During the annotation, each time we send a small "*.txt" file containing hundreds of dialogue examples to the annotators and conduct a random sampling test after they return the annotated data\footnote{During annotation, we randomly selected 5\% of the examples from one annotated file and checked if the annotator made accurate annotations for these random examples. The annotators were preliminary trained so that they were expected to make as few errors as possible. We expected no more than 1 error per 20 examples in the random sampling test. Otherwise, the file will be sent back for revision.}. The annotator who returns a low-quality file will be asked to check their annotation again before we send the next file. The whole annotation takes 35 days, and each dialogue is annotated by 3 annotators. When determining the final result, the majority will be adopted when there is a disagreement among the three annotators\footnote{There are a few cases where the three annotators disagree with each other, we decide these cases by ourselves.}. The overall inter-rater agreement measured by Fliess’ Kappa is $0.61$, indicating a substantial agreement among the annotators.


\begin{table}
\footnotesize
\centering
\begin{tabular}{l | c | c}
\hline
         \textbf{Category}     &  \textbf{Ch}  &  \textbf{En}  \\
\hline
Simile                      & 5,515  &  3,576 \\
Literal                      & 5,904  &  4,570 \\
\hline
\textbf{Tenor} in context        &  32.8\% & 48.9\%   \\
\textbf{Tenor} in response      &  67.2\% &  51.1\%   \\%
\hline
\textit{Vehicle} before \textbf{Tenor}       & 5.7\%  &  0.9\%   \\
\textbf{Tenor} before \textit{Vehicle}      & 94.3\%  & 99.1\%   \\
\hline
Ave. context words in simile & 20.76  &  22.22 \\
Ave. response words in simile& 18.86  & 17.83 \\
\hline
\end{tabular}
\caption{The statistics of the MSD dataset. "diff." means "different". "Ave." is short for "Average".}
\vspace{-0.5cm}
\label{MSD-statistics}
\end{table}

\subsection{Data Statistics}
After the annotation, we get a total of 19,565 (8,146 English and 11,419 Chinese) dialogues. The MSD has multiple comparators for both English and Chinese data. In MSD English data, the "like" mode is around 52.4\% and the "as" mode is around 47.6\%. In MSD Chinese data, "\begin{CJK}{UTF8}{gbsn}像...一样\end{CJK}" accounts for the most\footnote{There are total of 11 comparators in Chinese data. Please refer to Appendix \ref{statistic-appendix} for more details.}. The proportion of each comparator is similar in simile and literal data. Table \ref{MSD-statistics} shows some of the statistics of the MSD data. Please refer to the data link for more details.

\section{Tasks and Results}
In this section, we introduce the 5 tasks defined with our MSD dataset. Including the definition of the task, the baselines, evaluation metrics, experimental results, and analysis. The implementation details are shown in the Appendix \ref{implementation}.

\subsection{Simile Recognition Task}
Following previous work \cite{DBLP:conf/emnlp/LiuHSFLH18,DBLP:conf/coling/LiLG22a}, we define simile recognition as a binary classification task where the model needs to distinguish whether an input sequence contains a simile. The input is a multi-turn dialogue and the output is True (simile) or False (literal).

\subsubsection{Baselines and Evaluation Metrics}
We use two baselines: 1) BERT is widely used and proven to be effective in classification tasks. We randomly split our MSD-En/Ch data into train/validation/test (8:1:1) sets and use the train set to fine-tune BERT. We use the output vector of the first input token <cls> of BERT to calculate the classification score for the input dialogue (see Appendix \ref{implementation}); 2) a large language model (ChatGLM\footnote{https://www.datalearner.com/ai-models/pretrained-models/ChatGLM-6B}). The input to ChatGLM is a concatenation of three parts: the definition of simile "A simile is a figure of speech that compares two different things via their shared properties."; a requirement "answer yes or no to this question: is the following dialogue example contains a simile?"; a simile dialogue examples such as in Table \ref{examples}. Then we calculate the results according to the prediction of the baselines. Following previous work \cite{DBLP:conf/emnlp/LiuHSFLH18}, we use Precision/Recall/F1 to measure the results.

\begin{table}
\footnotesize
\centering
\begin{tabular}{l | c | c | c}
\hline
\textbf{Model}    & \textbf{Precision} & \textbf{Recall} & \textbf{F1}   \\
\hline
\multicolumn{4}{c}{\textit{MSD-En}}\\
\hline
ChatGLM(zero-shot) &0.4793	& 0.8441 & 0.6114 \\
BERT(fine-tuned)       & 0.7154 & 0.6759 & 0.6951 \\
\hline
\multicolumn{4}{c}{\textit{MSD-Ch}}\\
\hline
ChatGLM(zero-shot) &0.4992 & 0.8772 & 0.6363\\
BERT(fine-tuned)      & 0.7754 & 0.7519 & 0.7635 \\
\hline
\end{tabular}
\caption{Simile recognition results.}
\vspace{-0.5cm}
\label{recognition}
\end{table}

\subsubsection{Results and Analysis}
Table \ref{recognition} shows the simile recognition results. We can see that BERT(fine-tuned) performs much better on Precision and F1 than ChatGLM on both MSD-En and MSD-Ch\footnote{For Chinese, we use https://huggingface.co/bert-base-chinese}. It is reasonable since the BERT models are fine-tuned on our training set. On the other hand, the ChatGLM is much better on Recall with a zero-shot setting. Overall, the classification results on both BERT and ChatGLM still have a lot of room to improve. Using syntactic structure information to locate simile components may help this task.

\subsection{Simile Interpretation/Generation Tasks}
Following the previous simile interpretation task \cite{DBLP:journals/taslp/SongGFLL21,DBLP:conf/acl/HeCLXX22} and simile generation task \cite{DBLP:journals/taslp/SongGFLL21}, we define Simile Interpretation/Generation (SI/SG) as a Multi-choice task with the "as...as" mode in our MSD-En\footnote{We did not conduct simile interpretation/generation on MSD-Ch in this paper since we did not annotate the shared property in Chinese data and we leave it for future work.} data (we test with 450 examples) since the shared property naturally exists in the comparator.

For \textbf{interpretation task}, we have a simile dialogue where the shared property between two "as"s is removed and replaced with a blank. The model needs to select a property from 4 choices (one correct answer and three distractors) for the blank. We construct the distractors with ConceptNet \cite{DBLP:conf/aaai/SpeerCH17}. In particular, we first use the tenor and some relations to find the related concept to the tenor and then use the HasProperty relation to find the distractors. Notice that for the examples where the tenor is a phrase of a sentence we could not find in ConceptNet, we use keywords (e.g. the subject of the sentence, the noun in the phrase) as the tenor to search ConceptNet. 

Similar to the simile interpretation task, we remove the vehicle in a simile dialogue and leave a blank for the \textbf{simile generation task}. The model needs to select a proper vehicle for this blank from 4 candidates (one correct answer and three distractors). We also construct the distractors with ConceptNet. We use the vehicle and certain relations in the ConceptNet to find the related concepts to the vehicle as the distractors. Notice that for the examples where the vehicle is a phrase or sentence that we could not find in ConceptNet, we use the vehicles from other dialogues in MSD dataset as the distractors. 

To ensure the distractors are true-negative, we randomly select 50 dialogue examples and manually check the quality of the distractors. We find that 92\% of the distractors are well selected and the rest 8\% are not as ideal as we expected but can still serve as distractors. More details about using ConceptNets are shown in Appendix \ref{conceptnet}.

\subsubsection{Baselines and Evaluation Metrics}
The first baseline is a BERT-large model which takes the whole dialogue with the shared property or the vehicle masked and predicts the masked words. The second baseline is the BERT-Probe \cite{DBLP:conf/acl/HeCLXX22} that fine-tunes BERT with the simile interpretation task. To compare both SI and SG tasks with this baseline, we further fine-tune the BERT-Probe model with the SG task using the data proposed by \citet{DBLP:conf/acl/HeCLXX22}. The third baseline is BERT-ANT \cite{DBLP:conf/acl/ChenCZPCZXCS22} which is trained with masked word prediction with metaphor data and can solve the Simile Interpretation and Generation tasks in a unified framework of simile triple completion. For example, when giving tenor=fireman and vehicle=bull, BERT-ANT can generate a list of words including the shared property like "strong" or "brave". All baselines are based on a BERT-large-uncased model. Since there are multiple masked words in our SI/SG experiments. We encode the predicted words and the candidates into dense vectors with a sentence-transformer (huggingface.co/sentence-transformers/all-MiniLM-L6-v2). Then we compute the cosine similarity between the predicted words and each of the candidates. The candidate with the highest similarity is chosen as the answer. We use Hit@1 to measure the accuracy. 


\begin{table}
\footnotesize
\centering
\begin{tabular}{l | c | c }
\hline
\textbf{Model}    & \textbf{Interpretation}  &  \textbf{Generation} \\
\hline
BERT-large               &  0.5603   & 0.2967 \\
BERT-Probe             &  0.5804   & 0.3375 \\
BERT-ANT               &  0.4621   & 0.3337 \\
\hline
\end{tabular}
\caption{Simile interpretation and generation results (Hit@1) on MSD-En.}
\vspace{-0.5cm}
\label{interpretation}
\end{table}

\subsubsection{Results and Analysis}
Table \ref{interpretation} shows the results of simile interpretation/generation tasks. We can see that BERT-Probe performs better than BERT-large in this task, showing that a model pre-trained on simile data can better align the simile components in an input sequence and predict the missing component, even though the training data is much different from our proposed data. The BERT-ANT performs similarly to the other two models on SG tasks but not as well at SI. It is because the training data of BERT-ANT is more of a metaphor data rather than simile data, a large portion of the metaphor data does not have shared properties. Hence, BERT-ANT is more powerful in connecting tenor and vehicle but is less powerful when predicting shared properties. Overall, the results on both simile interpretations/generations still have a lot of room to improve. How to exploit the semantic information in context to help these tasks requires further study.

\subsection{Response Retrieval Task}
Following previous work in retrieval \cite{DBLP:conf/cikm/GuoFAC16}, we define Response Retrieval as a ranking task. The input is a multi-turn dialogue context and multiple response candidates (including the correct one) and the model needs to rank all the candidates so that the correct one has the highest score. In particular, for each "dialogue context" in MSD simile data (both English and Chinese), we randomly select 19 responses from other dialogue as the negative examples. 

\subsubsection{Baselines and Evaluation Metrics}
We use BERT-base for our baseline in response retrieval since it is widely used and proven to be effective in retrieval tasks. We concatenate dialogue context and each of the response candidates as the input sequence to the pre-trained model. Then we use the output of the first input token <cls> to compute the score for the input sequence as in Appendix \ref{implementation}. Finally, the response candidate with the highest score will be chosen as the answer. 

We first randomly split the Reddit dialogue data into train/validation/test (14.99M/5K/5K) sets. Then we used the BERT model to train an English dialogue retrieval model with this train/validation data. The model is denoted by BERT(Reddit). We choose a checkpoint with the best performance on the validation set. Then we use this checkpoint to compare its performance on both the Reddit Test set and the MSD-En set. Similarly, we combine LCCC and PchatbotW and randomly select 12M/5K/5K from the combined data as train/validation/test sets and train a Chinese dialogue retrieval model. The trained BERT\footnote{https://huggingface.co/bert-base-chinese} model is denoted by BERT(Ch) and used to do the comparison of the performance on the LCCC+PchatbotW Test set and the MSD-Ch set. We measure the accuracy of the retrieval with Recall@1/2/5. 

\begin{table}
\footnotesize
\centering
\begin{tabular}{l | c | c | c}
\hline
\textbf{Model}    & $\textbf{R}_{20}$@1 & $\textbf{R}_{20}$@2 &  $\textbf{R}_{20}$@5  \\
\hline
\multicolumn{4}{c}{\textit{MSD-En simile data}}\\
\hline
BERT(Reddit)        &  0.4212 & 0.4960 & 0.6391  \\
\hline
\multicolumn{4}{c}{\textit{Reddit Test set (5K)}}\\
\hline
BERT(Reddit)        &  0.8012 & 0.9066 & 0.9319  \\
\hline
\multicolumn{4}{c}{\textit{MSD-Ch simile data}}\\
\hline
BERT(Ch)        &  0.3706 & 0.4632 & 0.6191  \\
\hline
\multicolumn{4}{c}{\textit{LCCC+PchatbotW Test set (5K)}}\\
\hline
BERT(Ch)        &  0.4221 &  0.5217 &  0.8024  \\
\hline
\end{tabular}
\caption{Response retrieval results.}
\vspace{-0.5cm}
\label{dialogue-retrieval}
\end{table}

\subsubsection{Results and Analysis}
Table \ref{dialogue-retrieval} shows the results of the response retrieval task. The performance of BERT(Reddit) and BERT(LCCC) on MSD is lower than their performance on Reddit and LCCC+PchatbotW Test sets, respectively. The results show that the data distribution in MSD is different from the data used to extract it and selecting a simile response is much harder than selecting a proper response. The low Recall results show that the dialogue retrieval task on MSD simile data needs further study. This requires a model that judges not only the relevance between context and response but also the plausibility of similes.

\begin{table*}[htbp]
\footnotesize
\centering
\begin{tabular}{l     |c|c|c|c|c}
\hline
\textbf{Model} & \textbf{PPL} & \textbf{BLEU}(1/2/3/4)(\%) & \textbf{ROUGE}(1/2/L)(\%) & \textbf{METEOR}(\%) & \textbf{Distinct}(1/2)(\%)    \\
\hline
\multicolumn{6}{c}{\textit{Reddit-dialogue Test set (En)}}\\
\hline
DialoGPT  & 236.74 &  0.01 / 0.00 / 0.00 / 0.00 & 2.05 / 0.00 /1.79 & 1.24 & 6.67 / 23.84 \\
GODEL   & 3.70 & 0.53 / 0.02 / 0.00 / 0.00 & 2.80 / 0.00 / 1.98 & 2.41 &6.54 / 36.01 \\

\hline
\multicolumn{6}{c}{\textit{MSD-En (simile data)}}   \\
\hline
DialoGPT  & 329.55 &  11.29 / 3.58 / 1.45 / 0.70 & 7.53 / 0.57 / 6.39 & 8.48 & 8.39 / 28.16 \\
GODEL   & 6.10 & 17.10 / 5.99 / 2.61 / 1.37 & 10.91 / 0.87 / 8.94 & 11.78 & 7.00 / 23.37 \\

\hline
\multicolumn{6}{c}{\textit{MSD-En (simile data) on Response Completion}}   \\
\hline
DialoGPT  & - &  17.29 / 8.50 / 5.24 / 3.35  & 23.71 / 5.13 / 23.04 & 12.85 & 14.64 / 43.51 \\

\hline
\multicolumn{6}{c}{\textit{LCCC+PchatbotW Test set (Ch)}}\\
\hline
CDialGPT(Ch) & 102.00 & 3.01 / 0.64 / 0.16 / 0.05 & 5.42 / 0.21 / 4.77 & 2.24 & 11.10 / 40.41 \\
GPT-2(Ch)        & 129.28 & 5.20 / 1.50 / 0.59 / 0.26 & 7.09 / 0.87 / 6.14 & 3.04 & 23.23 / 66.14 \\
\hline
\multicolumn{6}{c}{\textit{MSD-Ch (simile data)}}  \\
\hline
CDialGPT(Ch)  &113.75& 3.07 / 0.72 / 0.26 / 0.09 & 5.46 / 0.24 / 4.85 & 2.30 & 11.36 / 40.58 \\
GPT-2(Ch)         &101.24& 5.89 / 1.11 / 0.27 / 0.10 & 6.35 / 0.19 / 5.47 & 2.98 & 12.15 / 48.18 \\
T5-base(Ch)      &118.60 & 7.61 / 2.57 / 1.40 / 0.94 & 8.66 / 0.94 / 7.66 & 4.25 & 22.15 / 66.59 \\
BART-large(Ch)&44.28&10.16 / 3.34 / 1.64 / 1.00 & 11.13 / 1.09 / 8.82 & 6.56 & 15.26 / 51.91 \\
\hline
\end{tabular}
\caption{Dialogue generation and completion results.}
\label{dialogue-generation}
\vspace{-0.5cm}
\end{table*}

\subsection{Response Generation Task}
The traditional response generation task uses dialogue context as input and outputs the response of the context. In this section, we also introduce a new generation task that completes the response sentence behind the comparator. Taking the fifth simile dialogue "Arguing with parents is not wise. It is like throwing an egg at a rock." as an example, we give the model "Arguing with parents is not wise. It is like" as input and ask the model to generate the rest "throwing an egg at a rock.". This is different from the Writing Polishment with Similes \citet{DBLP:conf/aaai/ZhangCXGLWC21} task since our task is a dialogue scene. The model needs to understand the difference between different speakers and complete the simile sentence. We use the simile data in MSD for the generation experiments. We conduct comparative experiments on the Reddit-dialogue Test set and the LCCC+PchatbotW Test set we used in the response retrieval task to show the difference between datasets.

\subsubsection{Baselines and Evaluation Metrics}
For the traditional response generation task, we use the DialoGPT \cite{DBLP:conf/acl/ZhangSGCBGGLD20} and GODEL \cite{DBLP:journals/corr/abs-2206-11309} for English data; use T5-base\footnote{huggingface.co/shibing624/prompt-t5-base-chinese}, BART-large\footnote{huggingface.co/HIT-TMG/dialogue-bart-large-chinese} \cite{DBLP:conf/acl/LewisLGGMLSZ20}, GPT-2\footnote{huggingface.co/shibing624/gpt2-dialogbot-base-chinese} \cite{radford2019language}, and CDialGPT\footnote{huggingface.co/thu-coai/CDial-GPT\_LCCC-large} \cite{DBLP:conf/nlpcc/WangKZHJZH20} for Chinese data. We choose these baselines since 1) they are widely used and proven to be effective in dialogue generation tasks. For example, GODEL (Grounded Open Dialogue Language Model) is pre-trained for dialogue and is initiated from T5 \cite{DBLP:journals/jmlr/RaffelSRLNMZLL20}. CDialGPT and BART-large are pre-trained with LCCC-large; 2) the different size models can provide more insight into the experiments. For our proposed response generation (completion) task, we conduct the experiment on English data with DialoGPT. 


We use the following automatic evaluation metrics employed by dialogue research. Perplexity (PPL), BLEU \cite{DBLP:conf/acl/PapineniRWZ02}, ROUGE \cite{lin2004rouge}, METEOR \cite{DBLP:conf/wmt/LavieA07}, and Distinct \cite{DBLP:conf/naacl/LiGBGD16}. PPL measures the probability of the model predicting the real response. BLEU measures the n-gram overlap between the generated response and the reference one. ROUGE is based on the calculation of the recall rate of the common sub-sequence of generating response and the real one. METEOR further considers the alignment between the generated and the real responses to improve BLEU. Distinct measures the diversity of responses by calculating the proportion of distinct n-grams in the total number of n-grams. Higher BLEU/ROUGE/METEOR/Distinct means better performance. The PPL is provided for comparing models with the same vocabularies, and the results are also useful for future research.

\subsubsection{Results and Analysis}
Table \ref{dialogue-generation} shows the generation and completion results. On most metrics of English data, DialoGPT and GODEL perform better on MSD-En than on Reddit-dialogue. CDialoGPT and GPT-2 have comparable performance on the LCCC+PchatbotW Test set and MSD-Ch. This is different from the response retrieval tasks where the MSD data is more difficult than the original data used to extract MSD. The reason may be the dialogue context in MSD provides more information than the context in the original data, so the generation models could leverage the rich context information to construct an informative response. Experiments also verify that larger models (GODEL/T5/BART) have a better performance. However, even the performance of the best baseline can still be improved. We analyze the generation results. Although there are some interesting cases, most of the results are not similes. It means the simile dialogue generation task requires a specific model design to capture the simile relations in context. We provide a case study in Appendix \ref{generation-cases}.

For the response completion task, when giving the comparator, DialoGPT has a big performance gain. It proves that the simile generation can benefit from the guide. Please refer to our code/data link for more experimental results about this simile dialogue completion task. 

\section{Conclusion}
We propose manually annotated multilingual simile dialogue (MSD) data for both simile and dialogue research. We design 3 simile tasks (recognition, interpretation, and generation) and 2 dialogue tasks (retrieval and generation) with MSD. Experiments with strong baselines show the challenge of each task. Future works include but are not limited to \textbf{1)} Dataset enlargement (e.g., more annotated examples with more kinds of comparators); \textbf{2)} Model designing (e.g., models with a specific structure to address the proposed tasks); \textbf{3)} New task designing (e.g., detecting tenor in the coarse/fine data). We encourage using the MSD in future simile and dialogue research.

\section*{Limitations}
Due to time constraints, we were unable to implement some unreleased models as baselines for the proposed tasks. We did not conduct simile interpretation/generation on MSD-Ch in this paper since we could not automatically annotate the shared property in Chinese data like the "as...as" mode in English. We are currently working on this annotation and plan to release the Chinese simile interpretation/generation results on the data link. The coarse/fine version data we introduced in this paper can still be used for enlarging the MSD data. We will study to utilize them for more simile data and richer language phenomena.

\section*{Ethics Statement}
We provide and emphasize some details of our work to address potential ethical concerns. First, all the data sources used in the data collection process are publicly available. We did not make any changes to the data sources and only extracted dialogue examples from these data. We carried out strict quality control during the extraction and annotation process. We made sure that there are no sensitive words even though the original data sources have already conducted this kind of checking. However, using our data to train or fine-tune a pre-trained generation model may still generate semantic errors or unpleasant similes or responses. One reason is that simile is a difficult task that compares two different things, mistakes could happen even when humans use similes. The other reason is that the knowledge stored in the original parameters of the pre-trained models may dominate the generation. We protect the privacy rights of annotators and paid 0.55 Chinese Yuan for annotating each dialogue data. The income of each annotator was above 100 Chinese Yuan per hour (On January 20, 2023, 100 yuan can be converted into 14.73 dollars).

\section*{Acknowledgements}
This paper is supported by the Science and Technology Innovation 2030 Major Project of China (No. 2021ZD0113302), the National Natural Science Foundation of China (No. 62076081, No. 61772153, and No. 61936010) and Nature Scientific Foundation of Heilongjiang Province(YQ2021F006).

\bibliography{anthology}
\bibliographystyle{acl_natbib}

\appendix

\section{Implementation Appendix}
\label{implementation}
The implementations of the pre-trained models in this paper are all based on the public Pytorch implementation \footnote{https://github.com/huggingface/transformers}. The hyper-parameters follow the default settings. We did not truncate any of the dialogue because the dialogue length in MSD data is much smaller than the maximum input length of the pre-trained models. We use a single Tesla v$100$s GPU with $32$gb memory to conduct experiments, the batch size is $8$ for all experiments. Checkpoints are chosen with the best performance on the corresponding validation set. In simile recognition and dialogue retrieval tasks, the first input position of the model $\mathcal{G}$ is a special token "<cls>", and the corresponding output vector $E_{cls}$ is fed into a non-linear layer to compute the final score of the input sequence:

\begin{align}
\mathcal{G}(input)\text{= } \sigma(W_2\cdot \mu(W_1\cdot E_{cls} \text{ + } b_1) \text{ + } b_2 ),
\end{align}

where $W_{1,2}$ and $b_{1,2}$ are training parameters; $\sigma/\mu$ is the sigmoid/tanh function, respectively. When training the simile recognition model, the loss is cross-entropy between predicted labels $y_i$ and ground-truth label $\bar{y}_i$: 

\begin{align}
\mathcal{L}_{simile} =  - \frac{1}{N} \sum_{i=1}^{N} (\bar{y}_i logP(y_i) )
\end{align}

Where $N$ is the number of simile examples. When training the dialogue retrieval model, the loss is calculated as follows:

\begin{align}
\mathcal{L}_{dr} \text{=} \text{ -} \sum_{i=1}^{N}\text{log}(\frac{e^{\mathcal{G}(\text{C}_i, \text{R}_i^+)}} {e^{\mathcal{G}(\text{C}_i, \text{R}_i^+)} \text{+} \sum_{j=1}^{\alpha} e^{\mathcal{G}(\text{C}_i, \text{R}_j^-)}}), 
\end{align}

where C is the dialogue context, R is the response, and $\alpha$ is a hyper-parameter meaning the number of different negative samples for a positive one. We set $\alpha$ = 9 in our training.



\section{Statistic Appendix}
\label{statistic-appendix}
In Table \ref{comparators-chinese}, we present all the comparators and their proportions in MSD-Chinese. 

\begin{table}[t]
\footnotesize
\centering
\begin{tabular}{l | c}
\hline
\textbf{Comparators} &\textbf{Proportion (\%)} \\
\hline
\begin{CJK}{UTF8}{gbsn}像...一样\end{CJK}  & 49.5 \\
\begin{CJK}{UTF8}{gbsn}跟...一样\end{CJK}  & 34.8\\
\begin{CJK}{UTF8}{gbsn}跟...似的\end{CJK}  & 11.6\\
\begin{CJK}{UTF8}{gbsn}像...似的\end{CJK}  & 2.7\\
\begin{CJK}{UTF8}{gbsn}像\end{CJK}  & 0.3\\
\begin{CJK}{UTF8}{gbsn}仿佛\end{CJK} & 0.3\\
\begin{CJK}{UTF8}{gbsn}简直是\end{CJK}  & 0.3\\
\begin{CJK}{UTF8}{gbsn}如...般\end{CJK}  & 0.2\\
\begin{CJK}{UTF8}{gbsn}像...般\end{CJK}  & 0.1\\
\begin{CJK}{UTF8}{gbsn}如...一样\end{CJK}  & 0.1\\
\begin{CJK}{UTF8}{gbsn}仿佛...一样\end{CJK}  & 0.1\\
\hline
\end{tabular}
\caption{Comparators in the Chinese MSD data.}
\label{comparators-chinese}
\end{table}

\begin{table}[t]
\footnotesize
\centering
\begin{tabular}{l}
\hline
\textbf{Relation:} \textit{Definition} \\
\hline
\textbf{RelatedTo:} \textit{The most general relation. There is some} \\
\ \ \ \   \textit{positive relationship between A and B, but ConceptNet}\\
\ \ \ \   \textit{can't determine what that relationship is based on the}\\
\ \ \ \   \textit{data. Symmetric. learn <-> erudition}\\
\textbf{Causes:} \textit{A and B are events, and it is typical for A to}\\
\ \ \ \   \textit{cause B. exercise -> sweat}\\
\textbf{Desires:} \textit{A is a conscious entity that typically wants}\\
\ \ \ \   \textit{B. Many assertions of this type use the appropriate }\\
\ \ \ \   \textit{language's word for "person" as A. person -> love}\\
\textbf{DistinctFrom:} \textit{A and B are distinct member of a set; }\\
\ \ \ \   \textit{something that is A is not B. Symmetric. red <-> blue; }\\
\ \ \ \   \textit{August <-> September}\\
\textbf{SymbolOf:} \textit{A symbolically represents B. red -> fervor}\\
\textbf{MannerOf:} \textit{A is a specific way to do B. Similar to "IsA",}\\
\ \ \ \   \textit{but for verbs. auction -> sale}\\
\textbf{LocatedNear:} \textit{A and B are typically found near each} \\
\ \ \ \   \textit{other. Symmetric.  chair <-> table}\\
\textbf{CausesDesire:} \textit{A makes someone want B.  having no food}\\
\ \ \ \   \textit{ -> go to a store}\\
\textbf{MadeOf:} \textit{A is made of B.  bottle -> plastic}\\
\hline
\end{tabular}
\caption{Relations in ConceptNet we used to find distractors. "<->" means Symmetric relation for A and B. "->" means Asymmetric relation that A entails B.}
\label{relations-conceptnet}
\vspace{-0.5cm}
\end{table}

\section{ConceptNet Appendix}
\label{conceptnet}
We use ConceptNet to construct the distractors in simile interpretation and generation tasks. ConceptNet is a knowledge graph that connects words and
phrases of natural language with labeled edges \cite{DBLP:conf/aaai/SpeerCH17}. Two concepts (A and B) are connected with relations such as "IsA" or "PartOf". In the \textit{simile interpretation} task, we need to find three distractors for the shared property of tenor and vehicle. We use the "Antonym" relation to extract the antonym of the property as the first distractor. We adopt the nine relations in Table \ref{relations-conceptnet} to find the related concepts to the tenor and then use the "HasProperty" relation to find the properties of these related concepts. Finally, we randomly choose two of the nine properties as the distractors. For example, if the (tenor, property, vehicle) is (fireman, strong, bull). We first have "weak" as the Antonym and the first distractor of "strong". Then we find that "fireman" is related to (RelateTo) "fire" and "fire" has a property (HasProperty) "hot". So "hot" is the second distractor for "strong". We can get up to nine distractors and choose two of them along with the Antonym as the final three distractors. Notice that for the examples where the tenor is a phrase or sentence that we could not find in ConceptNet, we use keywords (e.g. the subject of the sentence, the noun in the phrase) as the tenor to do the search in ConceptNet. In the \textit{simile generation} task, we also use the relations in Table \ref{relations-conceptnet} to find the related concepts to the vehicle as the distractors. Different from the interpretation task, for the examples where the vehicle is a phrase or sentence that we could not find in ConceptNet, we use the vehicles from other dialogues in MSD dataset as the distractors.

\begin{table}[htbp]
\footnotesize
\centering
\begin{tabular}{l}
\hline
\textbf{English example in MSD}\\
\hline
\textbf{Dialogue context:} \\
\textit{You'll never love someone exactly the same as her, } \\
\textit{because there will never be another person exactly }\\
\textit{like her. But, that doesn't mean you won't ever be as}\\
\textit{in love with someone else.}\\
\hline
\textbf{Ground truth response:} \\
\textit{It will just be a different kind of love, as unique to that}\\
\textit{person as a fingerprint, yet just as strong and deep as }\\ 
\textit{before, or maybe even stronger and deeper.}\\
\hline
\textbf{Response from DialoGPT:} \\
\textit{I know, I know. I just don't want to be like her.}\\
\textbf{Completion from DialoGPT} (additionally input "it will \\
just be a different kind of love, as unique to that person as"): \\
\textit{a gift which will never be forgotten.}\\
\textbf{Response from GODEL:} \\
\textit{I'm not sure if this is true, but I'm pretty sure I've been}\\
\textit{in love with someone exactly the same as her.}\\
\hline
\textbf{Chinese example in MSD}\\
\hline
\textbf{Dialogue context:} \textit{\begin{CJK}{UTF8}{gbsn}鲜艳的月季花，伴你快乐的生活。\end{CJK}}\\
\textit{(Bright Chinese rose flowers accompany your happy life.)} \\
\hline
\textbf{Ground truth response:} \textit{\begin{CJK}{UTF8}{gbsn}愿我们的生活像花一样美好！\end{CJK}}\\
\textit{(May our life be as beautiful as flowers!)}\\
\hline
\textbf{Response from CDialGPT:} \textit{\begin{CJK}{UTF8}{gbsn}谢谢\end{CJK}(Thanks!)}\\
\textbf{Response from GPT-2:} \textit{\begin{CJK}{UTF8}{gbsn}谢谢姐姐\end{CJK}(Thanks. Sister!)}\\
\textbf{Response from T5-base:} \textit{\begin{CJK}{UTF8}{gbsn}花攒锦簇\end{CJK}(The flowers gather}\\
\textit{in abundance.)}\\
\textbf{Response from BART-large:} \textit{\begin{CJK}{UTF8}{gbsn}月季花开的时候，我们\end{CJK}}\\
\textit{\begin{CJK}{UTF8}{gbsn}的心情也会像花一样美丽！\end{CJK}(When the Chinese rose }\\
\textit{opens, our mood will be as beautiful as the flower!)}\\
\hline
\end{tabular}
\caption{Case study of simile response generation task.}
\label{case-study}
\vspace{-0.5cm}
\end{table}

\section{Generation Cases Appendix}
\label{generation-cases}
As we introduced in the simile response generation section, most of the generated results are not similes since the baselines are not designed for this task. In Table \ref{case-study}, we provide two cases to show the dialogue cases in MSD and the generation results from different models. In the first English example, both DialoGPT and GODEL generate fluent responses and contain the comparator "like" or "as". However, both models fail to generate a simile response like the ground truth one. The Chinese example is extracted from the LCCC data, we can see BART-large performs the best and gives an informative response with a simile in it. The GPT-2 gives a general response and T5-base gives an informative response. The CDialGPT also gives a general response even if it is trained with the LCCC dataset. The two cases in Table \ref{case-study} further verify that simile dialogue generation is challenging. However, in the response completion task, when adding the comparator in the input, we can see the DialoGPT outputs a simile and makes the dialogue more vivid and interesting.

\end{document}